# Autonomous generation of different courses of action in mechanized combat operations

Johan Schubert, Patrik Hansen, Pontus Hörling, Ronnie Johansson
*Swedish Defence Research Agency*
*SE-164 90  Stockholm, Sweden*
{johan.schubert, patrik.hansen, pontus.hoerling, ronnie.johansson}@foi.se

**Abstract**

In this paper, we propose a methodology designed to support decision-making during the execution phase of military ground combat operations, with a focus on one's actions. This methodology generates and evaluates recommendations for various courses of action for a mechanized battalion, commencing with an initial set assessed by their anticipated outcomes. It systematically produces thousands of individual action alternatives, followed by evaluations aimed at identifying alternative courses of action with superior outcomes. These alternatives are appraised in light of the opponent's status and actions, considering unit composition, force ratios, types of offense and defense, and anticipated advance rates. Field manuals evaluate battle outcomes and advancement rates. The processes of generation and evaluation work concurrently, yielding a variety of alternative courses of action. This approach facilitates the management of new course generation based on previously evaluated actions. As the combat unfolds and conditions evolve, revised courses of action are formulated for the decision-maker within a sequential decision-making framework.

## 1 INTRODUCTION

In this paper, we develop a method for decision support during the execution phase of military ground combat operations, specifically focused on the actions taken. The technique generates and evaluates recommendations for alternative actions for a mechanized battalion, providing direct support for those actions. The method begins with an initial set of courses of action, which are evaluated based on their likely outcomes. Thousands of action options are systematically generated and assessed through a technique that aims to find progressively better alternatives with improved outcomes.

In the initial information situation before an operation and continuously throughout the operation, the methods provide valuable recommendations for one's actions. The goal is to confront and outmaneuver an equal or stronger opponent by effectively utilizing one's resources amidst multiple possible action alternatives.

The method generates action alternatives that are evaluated against the opponent's condition and actions, taking into account unit composition, strength ratios, types of attack and defense, and expected rates of advance. The assessment of the combat's outcome and the rate of advance is performed using field manuals grounded in historical experience. The generation and evaluation processes occur in parallel as action alternatives are produced. This allows for controlling the generation process for new action alternatives by using those that have already been assessed. As the battle progresses and conditions shift, updated action alternatives are generated for the decision-maker in a sequential decision-making process.

In Section 2, we present the problem statement, followed by a description of the scenario analyzed in Section 3. Section 4 details our method for generating configurations. In Section 5, we explain our event-driven simulation approach. Section 6 outlines the calculation of outcomes using the box method, while Section 7 illustrates how to determine mission outcomes for a specific configuration of blue forces. Section 8 discusses the implementation, and Section 9 introduces clustering as a technique for grouping similar configurations based on their structure and outcomes. In Section 10, we offer insights into detailed decision support through analysis of incremental simulation outcomes for selected configurations. Finally, Section 11 presents our conclusions.

## 2 PROBLEM STATEMENT

The question is how to most effectively position a mechanized battalion or part of it for an operation. We model the problem using a knowledge representation approach known as the box method, following military field manuals [1–3]. The box method entails a detailed analysis of a critical event, such as an area of engagement. When employing this method, the staff isolates a specific location and focuses on the vital events occurring within it. We create a digital representation of our side's sequential actions, which serves as the framework for generating alternative actions. We can vary the number of platoons on our side from 1 to 16, which may consist of different types. Based on the starting position, we aim to relocate these units to new positions, considering how to establish an effective position for the upcoming battle. We test the methodology using a scenario on Rådmansö, outside the city of Norrtälje in mid-Sweden, as shown in Figure 1.

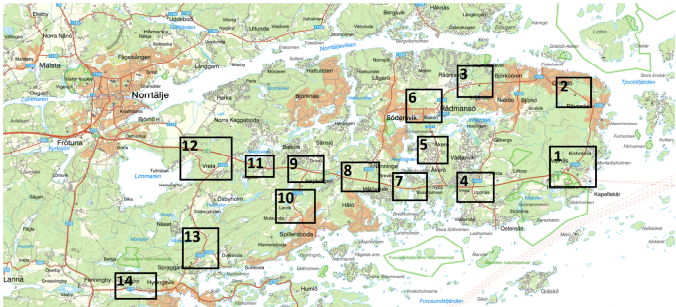

Figure 1: Boxes on Rådmansö (experiment with the box method).

We assess the outcome of the sequence of combats that occurs when the red forces advance west from box 1, considering alternative placements of the available platoons. A placement that results in a minimal reduction in *combat value* for the blue side and a significant reduction in combat value for the red side serves as a potential recommendation for the decision-maker. To effectively manage configurations with similar placements and comparable outcomes, we can categorize them into clusters, where different clusters are presented as aggregated results of various courses of action [4, 5].

## 3  SCENARIO

The movement patterns and positions of the red side were derived from the *Information Fusion Demonstrator 2003* (IFD03) [6, 7], which was presented to the Swedish Armed Forces in December 2003. This demonstration aimed to simulate an intelligence scenario in which an enemy mechanized battalion advances along various roads with multiple combat vehicles across Rådmansö. Different platoons can take alternative routes. No combat occurs in this scenario; instead, simulated information collection is conducted using fixed and mobile sensors. Observations from several geographically dispersed sensors are combined into a coherent situational picture, where algorithms utilizing known *ORBAT* templates consolidate the observed vehicles into higher-level units, such as platoons and companies. *Ground Truth* for red vehicle movement patterns was recorded during the simulation, and these records now serve as the foundation for describing the positions of the red platoons along the timeline.

To simplify, we have aligned the vehicle positions of the red platoon leaders in the IFD03 scenario with their respective locations in that scenario. Boxes have been placed throughout the scenario, centered at 14 locations along the Reds' various advance routes, as shown in Figure 2. The time for each red platoon to move from the position nearest to the center of a box to the corresponding position in the following box has been calculated using the Ground Truth logs. Red platoons bypass boxes without blue platoons; however, if one or more blue platoons are present when a red platoon arrives, combat ensues. The battle continues until one side is defeated. The red platoons follow a fixed advance strategy from one box to another, while the blue platoons experiment with different advance strategies. This means that red platoons advance along the designated geographical route, regardless of how the combat unfolds in the various boxes, as long as they are not defeated. Conversely, *different* red platoons advance along alternative routes.

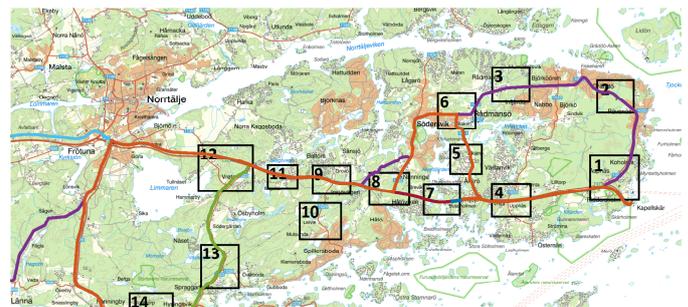

Figure 2: Fixed advance paths along the different routes.

## 4  GENERATION OF CONFIGURATIONS

To provide decision support in a situation involving a specific blue unit, we allow the decision methodology to autonomously generate and evaluate alternative configurations of initial positioning (starting groupings) for all available platoons. We assume that the blue side has some units available and wants to understand how to use them most effectively. Initially, we search for the most effective initial starting groupings, which are then dynamically simulated during combat against a red force across various scenarios. Subsequently, the groupings are evaluated based on the outcome (see Sections 5–6).

The number of alternative groupings depends on the total number of platoons and the variety of platoon types. With 14 boxes and 16 blue platoons, there are up to $14^{16} = 2.17 \cdot 10^{18}$ alternative groupings if all platoons are considered unique. Initial starting groupings must be generated and evaluated successively in rounds, after which new initial starting groupings can be created, taking into account the results of the evaluations conducted.

Generating configurations can be achieved using search and machine learning. For these methods, we select 256 pre-generated configurations, determined by a non-stochastic *Nearly Orthogonal Latin Hypercube* [8] (NOLH), which is represented as a matrix with 16 columns and 256 rows (S-NOLH(16, 256)). NOLH is a statistical selection method for computer experiments that ensures the input data are statistically independent and encompass the entire input space. Each column in the matrix corresponds to a potential platoon, while each row represents a configuration to be evaluated. If the number of platoons is less than the number of columns, one can choose as many columns as there are platoons. Any column can be selected (but no column more than once). The first 256 rows signify the initial configurations to be evaluated; thereafter, additional configurations can be generated. These initial 256 configurations serve as seeds for generating additional configurations.

Two methods for generating additional starting groupings involve processing batches of configurations sequentially, one batch at a time, through search and machine learning. Using these methods, we can create new starting groupings by effectively refining previously evaluated configurations, thereby reducing the number of groupings to assess to about 4500 in the case of ten platoons. We generate multiple configurations in a batch because we employ parallel programming. When evaluating a new configuration based on simulated results, we cannot use these results to create other configurations within the same batch. Consequently, new configurations are generated in batches, followed by subsequent evaluations.

As a search method, we employ *rank-order selection* [9], in which we randomly select a configuration based on the ranking of values for all configurations. The configuration with the lowest value is assigned a selection probability of 1/32896, while the next lowest receives a probability of 2/32896, and so on, up to a probability of 256/32896 for the configuration with the highest value. (Here, 32 896 is the sum of the numbers 1 to 256.) For the selected configuration, we then draw a random platoon with uniform probability. A new configuration is generated where the selected platoon is placed in a different box from its original location, with a ranking probability determined by the geographical distance to the other boxes. The box that is furthest away has a probability of 1/91, the one with the second furthest distance has a probability of 2/91, and so on, with a probability of 13/91 for the nearest box (the original box is assigned a probability of 0). (Here, 91 is the sum of the numbers 1 to 13.) The remaining platoons in the selected configuration remain unchanged in the new configuration. If the new configuration is unique and does not already exist among the 256 existing configurations, it is simulated. After simulation and evaluation, the new configuration is added to the group of the other 256 configurations, and the worst configuration among the 257 is eliminated.

We have also employed *genetic algorithms* (GA) [10] to generate new alternative configurations. In GA, we alternately use two operators: *mutation* and *crossover*. With a 5% probability, we select a mutation, randomly choosing one of the 256 configurations using rank-order selection. Then, with uniform probability, we choose a random platoon and, for this platoon, a new random box with a uniform probability of 1/13. With a 95% probability, we instead select crossover as the operator. Here, we draw two different configurations using rank-order selection. We create a new configuration by choosing a box with a 50% chance for each platoon from either the drawn configuration or the alternative. If the new configuration is unique, it is added to the group of the other 256 configurations after simulation and evaluation, and the worst configuration among the 257 is eliminated.

We can combine the two methods by using a ranked selection search with probability $p$ and a GA with a probability of $1 - p$. Studies in Chapter 8 indicate that the appropriate value falls within the interval $p \in [0.0, 0.6]$. In subsequent experiments, we use $p = 0.4$.

## 5 SIMULATING A CONFIGURATION

To evaluate all given configurations, they are simulated using the scenario. The initial situational state is established based on the specifications provided for the scenario. Within this state, the forces of both sides, unit positions, geography, and upcoming events are represented. The forces consist of a group of units, each identified by a specific unit type used to calculate its combat value. Additionally, all units are assigned a *relative combat value* that reflects the remaining portion of the original combat value after engagement. Units are designated as belonging to either the blue or red side, corresponding to one's own or the opponent's combat affiliation.

In the simulation, all units are positioned within the marked boxes. Given a map with deployed boxes, a mathematical representation of geography is derived by constructing a graph. Each node in the graph represents a box where potential combat could occur, while each edge indicates possible movement between two adjacent areas. When moving between boxes, units traverse along the connecting edge. Node properties include the coordinates of the box's center and its dimensions. The length of an edge is determined by calculating the road distance between the coordinates of the box centers using a mapping tool. For positions not represented in the graph that are not relevant destinations for blue unit movements but remain of interest, such as initial and final simulation positions, potential paths from these positions to nodes in the graph are incorporated. These positions are intended to facilitate one-way movements that direct units into or out of the graph.

When assessing a blue unit configuration, we simulate the movement of both blue and red forces between boxes and the ensuing battles within those boxes. The outcome table shows the results of each battle in every box, based on the relative combat values at the start of the battle. We disregard factors like terrain and weather.

In a simulation of the given scenario's configuration, all friendly units start in the initial position of Görla outside the graph. In contrast, the opponent's platoons begin in box number 1, as shown in Figure 2. The simulation operates on an event-driven model, meaning that updates to the situational state occur in response to specific events. Within this state, there is an event queue that lists all forthcoming events along with relevant information. The element in the event queue with the nearest start time is the next event to be executed. After an event has been executed, it is discarded, and the process continues with the next event in the queue.

There are two types of events in the simulation: *move-to-a-new-box* and *end-of-combat*. The simulation begins by creating a move-to-a-new-box event for each unit. The red platoons move according to the scenario, following a specified path with designated arrival times at each box. For a configuration of blue units, the platoons move toward a specific destination. Based on the unit's location and its destination, the shortest path is calculated through the graph as a sequence of movements between boxes along the route to the destination. An example of a state after executing the blue movements is shown in Figure 3. Travel time for a movement is calculated by considering the unit's speed and the distance covered. The speed of all blue units is supposed to be 30 km/h.

In a given configuration of blue units, all boxes are permissible destinations. During a blue unit's movement, it may encounter enemy units before reaching its destination. This occurs if the unit encounters a red platoon on an edge or is drawn into combat before reaching its intended destination. If this happens, it is deemed an illegal movement, resulting in a severe penalty during the evaluation, and this unit is discarded.

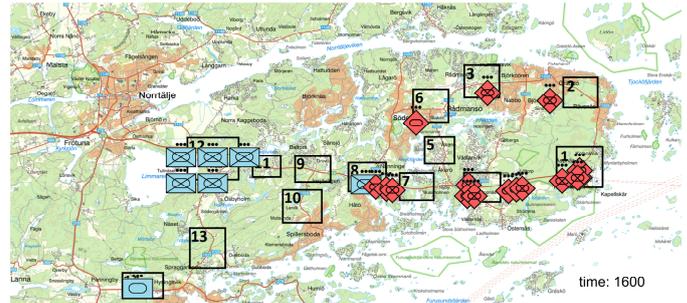

*Figure 3: Geographic representation of a state following the movements of one's platoons alongside the opponent's movements as outlined in the simulation scenario.*

When a move-to-a-new-box event is triggered, the affected unit is transferred to that location. Three different scenarios can occur when a unit arrives at a box: no enemy units are present, a new combat is initiated, or an ongoing combat in the box is interrupted. If a red unit arrives at a box and encounters no blue units, the platoon continues along its path to the next box. Conversely, if a blue unit arrives without encountering any enemy units, it remains in the box, waiting for the enemy to arrive. When a unit arrives at a box containing enemy units, a new combat begins, and an end-of-combat event is added to the event queue when one side is deemed defeated. If an ongoing combat is in progress in the box, an end-of-combat event occurs when new units arrive, followed by a temporary outcome calculated based on the combat that has occurred up to that point. The combat then resumes with the newly arrived units, and a new end-of-combat event is added to the event queue.

At the end of the end-of-combat event, the relative combat values of all participating units are updated. The outcome of the event is calculated in accordance with Section 6. If a unit's relative combat value drops to zero, it is considered eliminated and removed from the simulation. If the red side triumphs in the combat, a move-to-a-new-box event is added to the event queue for each platoon according to the scenario. If the blue force emerges victorious, multiple alternative configurations for the remaining units are evaluated, and the units are assigned to different boxes based on the evaluation. The number of these configurations is limited to a maximum of 40 times the number of remaining units exiting the box, to control computational costs. In each alternative, units are randomly divided into groups of varying sizes. The size of the first group is sampled from a discrete uniform distribution between 1 and the number of remaining units at the end of combat. (In the case of one remaining unit, there will only be 13 alternatives, one less than the number of boxes.) The next group is formed similarly, with the remaining units not assigned earlier, and so on. A new configuration is generated, containing a random destination for each group of blue units. All alternative event developments create their situational states through various action alternatives and combat outcomes. This methodology enables a consequence analysis of the configuration, evaluating potential future actions. Once all events are completed and the event queue is empty, the final state of the action alternative is assessed in Section 7. When multiple different actions have been explored following combat, the simulation pursues the movements of blue units that yield the best evaluation. After all options have been explored and the initial event queue is empty, the simulation concludes, and the assessment of the final state is assigned to the explored configuration.

# 6 COMBAT IN A BOX

## 6.1 OUTCOME OF THE COMBAT

The outcome of a combat in a box depends on the forces' initial combat values, the number of units, unit types, and the relative combat values, which are updated based on previous combats that the units have participated in. It also takes into account which side is attacking or defending, whether there is a meeting engagement, and the types of attack and defense. The duration of the combat is primarily determined by the nature of the combat, along with the unit type and combat values.

We base the combat outcome on the combat power analysis outcome tables from the U.S. Army School at Fort Leavenworth. Initially, all unit types are assigned a combat value where 1.0 corresponds to a fully capable armored battalion. Smaller unit types receive lower values, while other unit types obtain initial combat values based on an assessment (Appendix: Combat value). If a unit is structured in a manner not found in the tables, its combat value is determined by summing the combat values of the constituent platoons. This methodology mirrors the one previously used in data farming within the NATO group MSG-124 [11–13] in collaboration with the Bundeswehr Office for Defense Planning.

The tables outline presumed losses in various typical scenarios where the relative combat values differ for each side. When the force ratios differ from those shown in the tables, interpolation is applied.

When losses are minor (according to the tables), it is assumed that combat will continue with updated values until one of the two forces is eliminated. The elimination level is parameterized and set to a relative combat value of 0.3 times the initial combat value. Units with values below 0.3 are eliminated, and the winning side becomes available for new tasks.

An initial combat value, $combatvblue$, is calculated for each blue platoon based on the unit type from Table 3 (Appendix: Combat value).

We have,

$$combatvblue = \sum_{x} \frac{x.combatv}{16 \cdot x.size} \cdot x.rel \quad (1)$$

where $x$ represents a platoon of type $x.type$ with explanatory text $x.text$. Here, $x.size = 1$ indicates that $x.combatv$ applies to the entire battalion (Table 3). All of these are fixed parameters. Additionally, we have $x.rel$, which represents a relative combat value variable that is initially set to 1.0 for all units and adjusted downward during combat.

For red platoons, we calculate the initial combat values based on the data presented in Table 4 (Appendix: Combat value).

We have,

$$combatvred = \sum_{y} \frac{y.combatv}{16 \cdot y.size} \cdot y.rel \quad (2)$$

where $y$ represents a platoon of type $y.type$, accompanied by explanatory text $y.text$. Here, $y.size = 1$ indicates that $y.combatv$ applies to the entire battalion (Table 4). All of these are fixed parameters. Additionally, we have $y.rel$, a relative combat value variable initially set to 1.0 for all units and adjusted downward during combat.

When combat occurs in a box, the relative combat values of all platoons (both blue and red) are updated according to

$$\left\{ x.rel := x.rel \cdot \frac{combatvblue}{combatvblue_{old}} \right\}_x \quad (3)$$

and

$$\left\{ y.rel := y.rel \cdot \frac{combatvred}{combatvred_{old}} \right\}_y \quad (4)$$

where $combatvblue_{old}$ and $combatvred_{old}$ signify the initial values of the units at the start of combat, and $combatvblue$ and $combatvred$ are the new combat values after the combat in the box has concluded.

The relative arrival times of both sides to the box influence the type of combat that takes place. In this context, the relative arrival time is measured against three parameters established by the user, which determine whether the combat type is classified as *Meeting Engagement* for both sides, *Deliberate Defense* and *Deliberate Attack* for the blue and red sides respectively, or *Hasty Defense* and *Hasty Attack* for the blue and red sides respectively, or vice versa.

We can calculate the losses for both the blue and red sides using Table 5 (Appendix: Losses), based on the type of combat and the initial strength ratio of $combatvblue_{old}$ and $combatvred_{old}$. New relative combat values are calculated according to equations (3–4). If the strength ratios deviate from the table, new combat values are interpolated.

If neither side receives a relative combat value, $x.rel$ or $y.rel$, below a threshold value of 0.3, the procedure is repeated until one side reaches exactly 0.3. At that point, this side is assigned a relative combat value of 0 and exits the simulation simultaneously, while the other side's relative combat value is updated according to the table.

## 6.2 TIME OF COMBAT

The duration of combat is calculated based on the types of units and their relative strength ratios. Table 6 (Appendix: Rate of advance) provides the rate of advance under various conditions, measured in kilometers per Day. The duration of combat (measured in seconds) is determined by dividing the distance (measured in meters) over which the combat occurs by the rate of advance (measured in meters per second, according to

$$st = \frac{boxlength}{advrate} \quad (5)$$

where $st$ represents the combat time measured in seconds, $boxlength$ is the square root of the box area measured in meters, and $advrate$ signifies the advance rate in meters per second (converted from the table value in Table 6).

## 7 VALUATION OF A SIMULATION

After completing the simulation, we calculate the remaining combat values for both sides.

We have,

$$combatvred_{final} = \sum_{y} \frac{y.combatv}{16 \cdot y.size} \cdot y.rel_{final}. \quad (6)$$

This is an estimate of the combat value of the entire red force that could break through the blue defense (if $combatvred_{final} > 0$). We conduct the same calculations for the blue side. Once the simulation concludes, we obtain

$$combatvblue_{final} = \sum_{y} \frac{x.combatv}{16 \cdot x.size} \cdot x.rel_{final}. \quad (7)$$

This assessment measures the combat effectiveness of the entire blue force after the simulation.

The outcomes of the combined battles determine the value, $X_{value}$, of a specific configuration for blue forces. We aim to minimize the number of red breakthroughs ($combatvred_{final}$) while also minimizing blue losses ($combatvblue_{initial} - combatvblue_{final}$) and maximizing red losses ($combatvred_{initial} - combatvred_{final}$) in that order of priority.

We select the configuration that minimizes $X_{value}$ across all configurations $X$, where

$$\begin{aligned} X_{value} &= combatvred_{final} \\ &+ \alpha \cdot (combatvblue_{initial} - combatvblue_{final}) \\ &- \beta \cdot (combatvred_{initial} - combatvred_{final}) \end{aligned} \quad (8)$$

where $\alpha = 0.2$ and $\beta = 0.1$.

Since the initial values remain constant across all configurations, we can simplify the formula to

$$\begin{aligned} X_{value} &= combatvred_{final} - \alpha \cdot combatvblue_{final} \\ &+ \beta \cdot combatvred_{final} \\ &= (1 + \beta) \cdot combatvred_{final} \\ &- \alpha \cdot combatvblue_{final}. \end{aligned} \quad (9)$$

The preferred configuration $X$ is the one that minimizes $X_{value}$.

## 8 IMPLEMENTATION AND RESULTS

The program developed is implemented in MATLAB using the *Parallel Computing Toolbox* for maximum efficiency. Based on the provided scenario parameters, the blue and red forces are instantiated together with a graph for geographical representation. These parameters, along with the additional scenario information, are integrated into the initial situational state upon which all simulations are based. The red force in the scenario comprises 13 armored infantry platoons from the *Infantry Battalion* (BMP-3), as listed in Table 4 (Appendix: Combat value), and three tank platoons from the *Independent Tank Battalion* (51xT80). The types of platoons included in the Blue Force are armored infantry and tank platoons, similar to those on the opposing side. The armored infantry platoons are considered to belong to the Infantry Battalion (M2) in Table 3 (Appendix: Combat value), while the tank platoons are part of the *Armor Battalion* (M1A1). When evaluating the model, the ratio of blue armored infantry platoons to tank platoons is used to align closely with the same ratio in the red force.

For a given number of blue platoons described by the NOLH, simulated outcomes are calculated. Based on the initial state of the scenario, the newly generated configurations are examined and evaluated. Each configuration is simulated in accordance with Section 5. In one iteration, a group of twelve configurations is explored and assessed in parallel. All unique configurations are evaluated and added to the group of 256 configurations. The same number of configurations with the lowest values is then discarded. The algorithm is considered to have converged when no new configuration is found among the 40 best-valued configurations over 17 consecutive iterations.

In generating new configurations in Section 4, a parameter $p_{method}$ is defined to establish the balance between the search and GA algorithms. To determine the value of $p_{method}$ that produces the most favorable outcomes, the case of ten platoons is examined. The evaluation of the best-discovered configuration for the ten platoons with varying values of $p_{method}$ is presented in Figure 4. For each value of $p_{method}$, the expected value ($X_{value}$) of the best found configuration is estimated from 100 simulations, along with the standard deviation of the mean for variance analysis. Based on the results, $p_{method} = 0.4$ yields the most favorable outcome, indicating that the Search method is applied with a frequency of 0.4 and GA with a frequency of 0.6. Therefore, we have decided that the remaining runs will be conducted with $p_{method} = 0.4$.

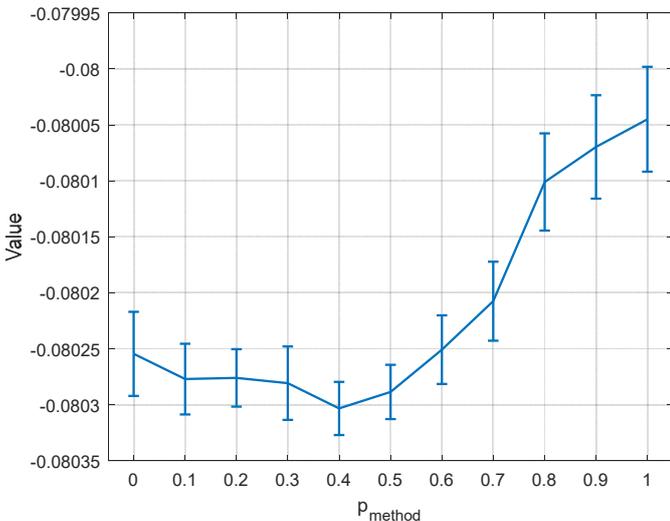

Figure 4: Estimation of the expected value from the simulations for the optimal configuration explored by the algorithm for ten platoons. For each value of $p_{method}$, 100 repetitions were performed to estimate the expected value. The standard deviation of the mean is provided with each estimate to assess variance.

To analyze the performance of various numbers of blue platoons, we estimate the expected value of the simulation outcome for the best-valued configuration based on ten repetitions for each number of platoons. The number of platoons ranges from 1 to 16, as shown in Figure 5. For each series of repetitions for a given number of platoons, the variance is displayed with the standard deviation of the mean. In this scenario, if the value of a configuration is lower than zero, the red force is halted, and the blue side emerges victorious. From this graph, it can be seen that the threshold for the number of blue platoons needed to stop the red force is seven platoons.

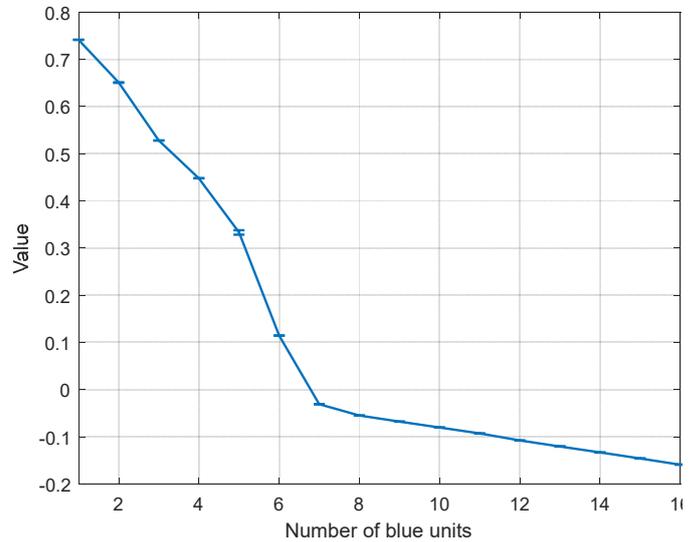

Figure 5: Estimation of the expected value for the best-valued configuration compared to the number of blue platoons across ten repetitions for each platoon count. For each number, the standard deviation of the mean is included to assess variance.

The time required for a simulation depends on the number of platoons in the blue force. For instance, with ten platoons, the expected elapsed time is estimated at 456 seconds, with a standard deviation of 85 seconds (when running in parallel on twelve CPU cores). To analyze the program's potential as an *anytime algorithm* [14], we examine how the value of the best-detected configuration evolves with each iteration. The mean of 100 repetitions is calculated for each iteration. In Figure 6, the results for seven and ten blue platoons are shown, along with the value of the best-detected configuration achieved thus far.

The value of the 256 best configurations from simulations for both seven and ten platoons is illustrated in Figure 7. One reason for the significant difference in the variance of the value during the earlier iterations for seven platoons compared to ten platoons, as seen in Figure 6, might be that only a few configurations result in the red force being halted, as shown in Figure 7. Consequently, in the initial iterations, it is less likely that a winning configuration will be discovered for seven platoons. In contrast, for ten platoons, the algorithm finds a winning configuration more rapidly.

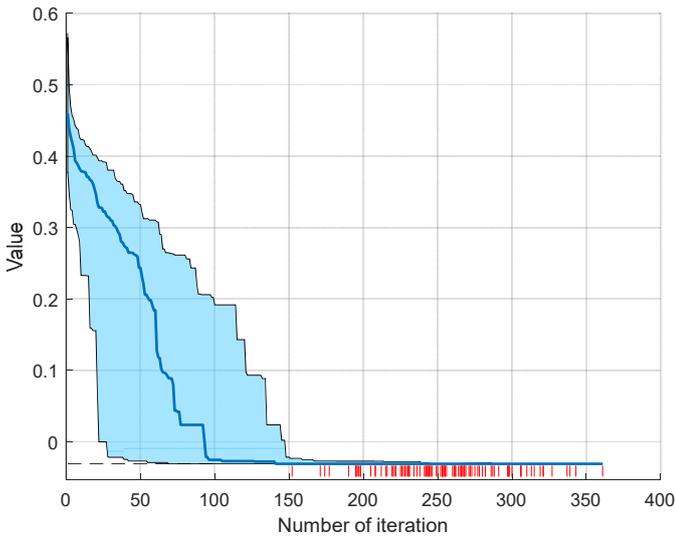

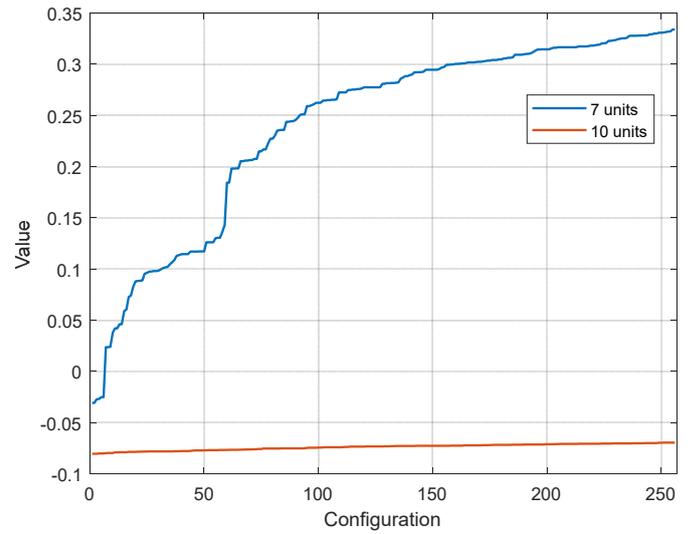

Figure 7: The values of the 256 best-explored configurations from simulations for seven and ten platoons.

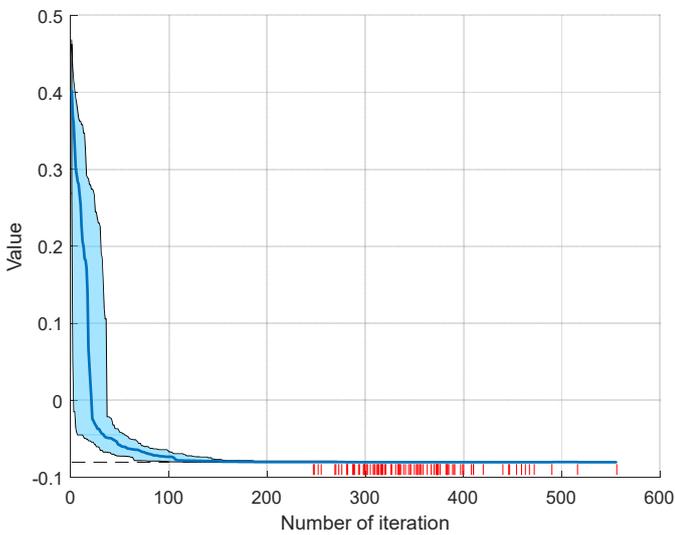

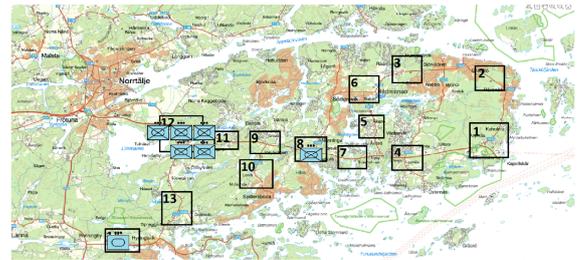

Figure 6: The blue curve represents the mean value of the best-discovered configuration for each iteration from 100 repetitions (top seven platoons; bottom ten platoons). The shaded area illustrates 95% of all data, excluding 2.5% of the lowest and highest values. The red vertical lines indicate the final iteration of a simulation, while the black horizontal dashed line shows the value of the best configuration across all iterations.

Figure 8 illustrates a graphical representation of the recommended course of action for seven platoons.

Figure 8: An example of a visual representation of the recommended movement for seven platoons. The abbreviation AIP nr refers to the number of the armored infantry platoons, while TP nr refers to the number for the tank platoons.

## 9 CLUSTERING OF EVALUATED CONFIGURATIONS

A decision support system can be designed to present only the best-generated configuration to the decision-maker based on the valuation in equation (9). However, this approach restricts the decision-maker's ability to use their expertise during the decision-making process. They are left with the choice of either accepting the system's suggested configuration or rejecting it in favor of an alternative. In the worst-case scenario, this could cause the decision-maker to ignore the decision support system altogether.

One way to give the decision-maker more flexibility is by showing a range of effective configurations that have been identified and evaluated. However, since the exploration process has generated and assessed thousands of configurations, many of the best options may look similar. This similarity can limit the decision-maker's ability to compare and choose freely among them.

To streamline the management of generated configurations, we can cluster similar ones and display only the best from each group.

The clustering algorithm starts by choosing a configuration from the set of generated configurations and placing it into a new cluster. It then evaluates another configuration and adds it to the same cluster if it is sufficiently similar to the first. If it isn't similar enough, a new cluster is created for that configuration. This process continues for all remaining configurations. Each new configuration is assessed to determine whether it is sufficiently similar to all the existing configurations in a cluster (and is then assigned to the best-fitting cluster) or if it should be placed in a new cluster.

After clustering all configurations, the configuration with the highest value in each cluster is identified. The average allocation of platoons to boxes within each cluster is also calculated, along with specific statistics for the cluster members.

We have identified two types of similarity: (i) purely structural similarity and (ii) similarity that involves both structure and value.

Structural similarity is defined solely by the degree of similarity between two configurations. For example, if two configurations place all platoons in the same boxes except for one, they receive a similarity score of one. The more deviations there are, the higher the score. This method allows us to group configurations with similar platoon placements and select only the best one from each group to present to the decision maker.

The second measure of similarity also looks at the values of the clustered configurations. In this case, clustered configurations need to share similar values and have a similar structure.

The second similarity measure is presented in equation (10). We have,

$$Similarity(X^i, X^j) =$$
$$1 - \left(1 - \frac{structSim(X^i, X^j)}{|X^i|}\right)$$
$$\cdot \left(1 - \frac{|X^i_{value} - X^j_{value}|}{max_k(X^k_{value}) - min_k(X^k_{value})}\right) \quad (10)$$

where $X$ represents the set of all evaluated configurations under consideration. Let $X^i$ and $X^j$ denote two configurations within the set $X$. The function $structSim(X^i, X^j)$ measures the structural difference between $X^i$ and $X^j$. The values of these configurations are represented by $X^i_{value}$ and $X^j_{value}$, respectively.

In equation (10), $structSim(X^i, X^j)$ is normalized by the number of blue platoons in the configurations, ensuring independence of the number of platoons. Analogously, the difference in configuration values is normalized by considering the range of configuration values, which is defined as the difference between the largest and smallest configuration values that have been calculated. When both the structural similarity and the differences in configuration values are minimized, $Similarity(X^i, X^j)$ is also minimized.

Using the first type of similarity often creates clusters that share similar structures, which makes sensitivity analysis easier (i.e., understanding if a slight change in a configuration can have significant effects). Conversely, the second type of similarity typically groups configurations with similar values, aiding in the extraction of features for improved decision-making (i.e., common properties among cluster members). Since clusters of the second type generally contain configurations with comparable values, decision-makers can identify high-quality configurations within these clusters. In our work, we utilize the second kind of similarity.

Figure 9 shows 51 clusters of configurations generated in a scenario with seven platoons (six AIP and one TP). The clusters are sorted based on the configuration value of their best member, with the lowest value considered the best. Error bars display the highest and lowest values within each cluster, with the median value indicated on each error bar. The dashed horizontal line at 0 marks the boundary between winning and losing configurations, with winning configurations located below the line and losing ones above. In this case, only two clusters include configurations that lead to victory.

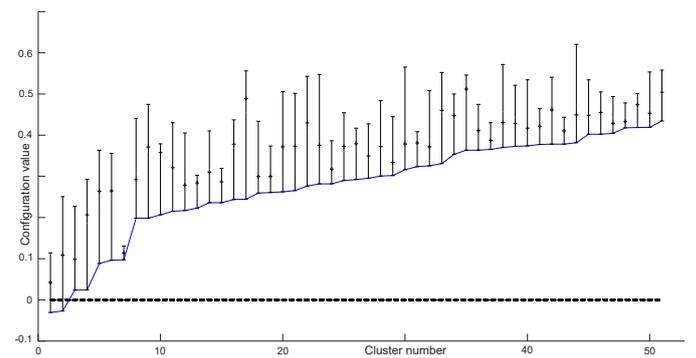

*Figure 9: Statistics of the 51 configuration clusters for the scenario involving seven platoons.*

In a scenario with ten platoons (eight AIP and two TP), the corresponding clustering is shown in Figure 10. This scenario has more clusters with winning configurations than the scenario with seven platoons.

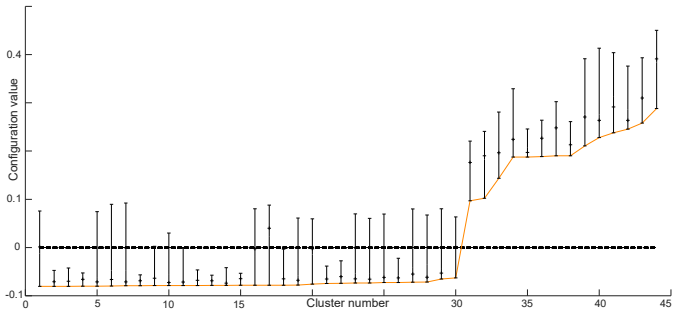

*Figure 10: Statistics for the 44 clusters of configurations in the scenario involving ten platoons.*

Figures 9 and 10 provide insights into the performance of the clustering algorithm, but may not be very helpful for decision-makers. Therefore, we show the ten best clusters from the previous analysis in Figures 11 and 12. Each cluster is depicted by a larger circle with a smaller circle inside. The larger circle displays a number indicating the cluster's rank. It is important to note that Cluster 1 represents the single best configuration. Additionally, the size of the larger circle corresponds to the number of members in that cluster.

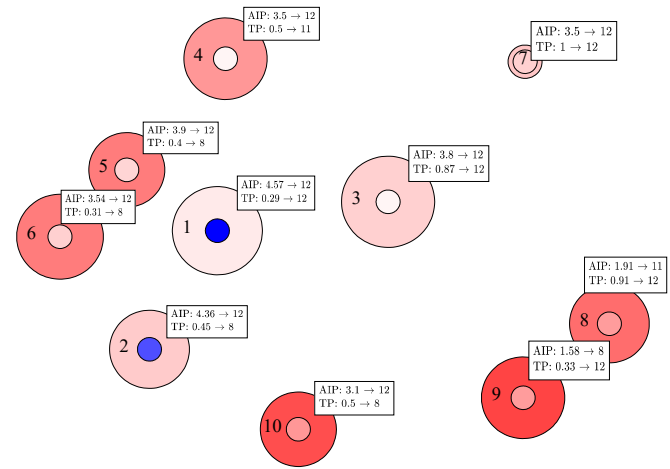

*Figure 11: Visualization of the top ten clusters for the scenario with seven platoons.*

The color of the small circle indicates the configuration value of the cluster's best member. Shades of blue represent negative values, suggesting that the blue side is winning in this scenario. A white circle signifies a configuration value of 0, while darker shades of red denote higher configuration values, with the red side prevailing. Each cluster includes a text box displaying the average number of AIP and TP platoons assigned to the most common geographical boxes, based on all configurations within the cluster. For example, in cluster 1 shown in Figure 11, an average of 4.57 AIP platoons and 0.29 TP platoons are assigned to box 12.

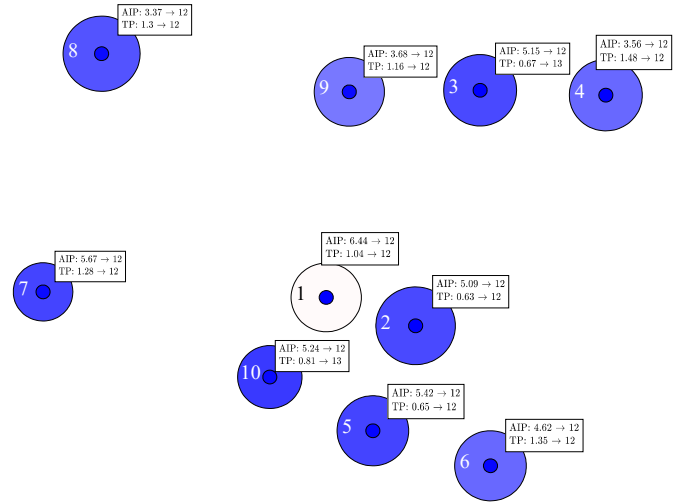

*Figure 12: Visualization of the top ten clusters for the scenario with ten platoons.*

The layout of the clusters in the visualization is designed to position similar clusters with the best members, as determined by equation (10), close to each other. This setup helps the decision-maker easily identify groups of configurations that share similarities in both structure and configuration value.

Upon examining cluster 1 in Figure 11, we find 14 alternative configurations where, on average, the platoons are distributed as shown in Table 1.

*Table 1: Average distribution of platoons among 14 configurations in cluster 1 for the case of seven platoons.*

| Average number of AIP | Box number | Average number of TP | Box number |
|---|---|---|---|
| 0.50 | 8 | 0.21 | 8 |
| 0.50 | 11 | 0.07 | 10 |
| 4.57 | 12 | 0.29 | 12 |
| 0.21 | 13 | 0.21 | 13 |
| 0.21 | 14 | 0.21 | 14 |

Table 2 provides a detailed breakdown of platoons arranged within the best configuration in cluster 1, offering an overview of the proposed resource allocation. This structured setup aims to significantly improve overall mission effectiveness by strategically positioning each platoon, ensuring that all available resources are utilized in the most efficient way to optimize operational outcomes.

*Table 2: Distribution of the seven platoons for the best configuration in cluster 1.*

| Number of AIP | Box number | Number of TP | Box number |
|---|---|---|---|
| 1 | 8 | 1 | 14 |
| 5 | 12 | | |

## 10 DECISION SUPPORT

To conduct a thorough analysis of the anticipated battles involving the blue forces and their potential outcomes, we will examine each engagement individually to identify key decisive moments. This meticulous approach enables decision-makers to better understand critical situations that could significantly impact the course of combat and focus their efforts on the most essential aspects. To conduct a comprehensive analysis of the anticipated battles and their possible outcomes involving the blue forces, we will scrutinize each engagement. By identifying pivotal moments within these conflicts, we can gain a deeper understanding of the dynamics at play.

This in-depth examination not only provides insights into the strategies employed but also equips decision-makers with a clearer understanding of critical situations, enabling them to focus their attention on the most significant factors influencing the course of the battles.

Figures 13–15 show a series of three simulated images that illustrate how combat values for both sides change over time, along with the number of remaining platoons in each force. In Figure 13, we see the initial deployment of our forces, arranged according to the best configuration in cluster 1 for this scenario, which features seven strategically positioned platoons. This layout is designed to maximize our defensive capabilities. Moving to Figure 14, we witness the first engagement. Here, a delaying battle unfolds in the terrain of Box 8, where tactical maneuvers are employed to slow the advancing enemy. Finally, Figure 15 showcases the culmination of the conflict in a decisive battle. In this crucial moment, our side takes a deliberate defensive stance and secures victory, showing the effectiveness of our planning and execution.

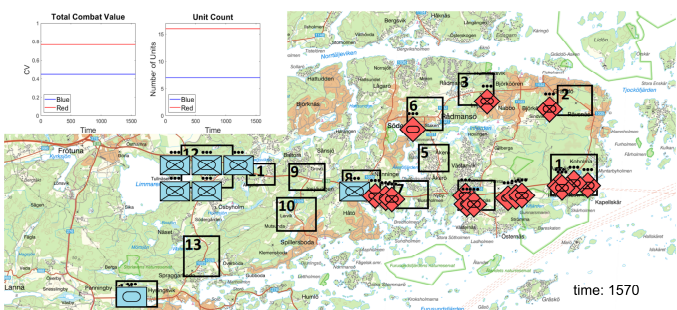

*Figure 13: The initial best blue configuration of seven platoons meets the first red forces in box 8.*

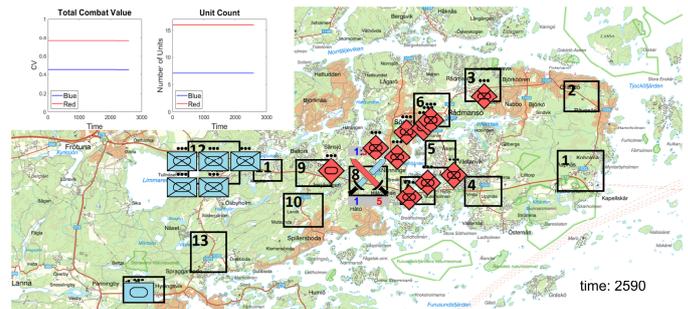

*Figure 14: An ongoing delaying battle occurs in box 8.*

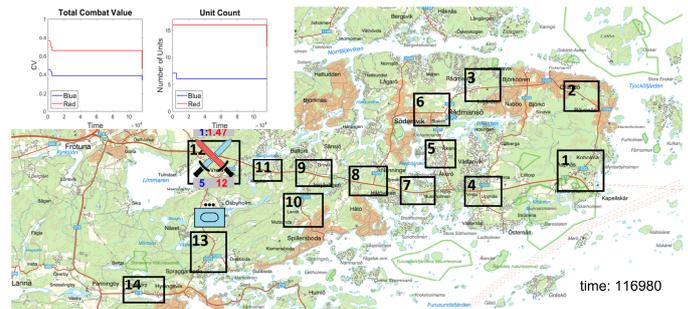

*Figure 15: The final combat in box 12 results in a victory for the blue forces.*

## 11 CONCLUSIONS

We have demonstrated that it is feasible to provide decision support during the execution phase of a military ground combat operation. This conclusion arises from a combination of qualitative and quantitative analyses of the outcomes. The computational speed enables the completion of a comprehensive analysis within a reasonable timeframe for decision-makers in certain scenarios, thanks to parallel processing. Nevertheless, in other instances, it may still be appropriate to present evaluations to the decision-maker sequentially via an anytime algorithm. This methodology is straightforward to implement, as the alternative generation approach systematically searches for improved starting configurations. Several research questions remain, including whether simulated results can be evaluated in an offline study using various methods, if partial outcomes can serve as seeds for new generations during dynamic replanning, the optimal techniques for grouping results being assessed to provide clear decision support (both conceptually and visually), and the nature of the interaction between the system and the decision-maker.

**APPENDIX: COMBAT VALUE**

Table 3 outlines the initial combat values for the blue side, where $x$ represents a platoon of type $x.type$, accompanied by the explanatory text $x.text$. Additionally, $x.size = 1$ indicates that $x.combatv$ applies to the entire battalion [12, 13]. All of these parameters are fixed.

*Table 3: Combat values for the blue side.*

| x.type | x.text | x.combatv | x.size |
|---|---|---|---|
| 1 | Infantry Bn (M113) | 0.71 | 1 |
| 2 | Infantry Bn (M2) | 1 | 1 |
| 3 | Infantry Bn (Light) | 0.48 | 1 |
| 4 | Infantry Bn (Airborne/Air Assault) | 0.7 | 1 |
| 5 | Separate Brigade (Armored) | 5.3 | 3 |
| 6 | Separate Brigade (Mech) | 4.7 | 3 |
| 7 | Separate Brigade (Light) | 3.1 | 3 |
| 8 | Armor Bn (M1A1) | 1.19 | 1 |
| 9 | Armor Bn (M1A2) | 1.3 | 1 |
| 10 | Armored Cav Regiment | 7.6 | 3 |
| 11 | Armored Cav Squadron | 2.2 | 1 |
| 12 | Regimental Aviation Squadron | 0.91 | 1 |
| 13 | Cav Troop (Ground) | 0.5 | 0.25 |
| 14 | 105(T) Bn (M102) | 0.8 | 1 |
| 15 | 105(T) Bn (M119) | 0.8 | 1 |
| 16 | 155(SP) Bn (M109A5) | 1 | 1 |
| 17 | 155(SP) Bn (M109A6) (Paladin) | 1.5 | 1 |

| 18 | 155(T) Bn (M198) | 0.8 | 1 |
|---|---|---|---|
| 19 | MLRS Bn | 4.5 | 1 |
| 20 | ATACMS Bn (B2) | 7.5 | 1 |
| 21 | ATACMS Bn (B1) | 8.8 | 1 |
| 22 | Div Cav Squadron (AASLT, Abn, Lt Div) | 0.7 | 1 |
| 23 | Div Cav Squadron (Heavy Div) | 3.8 | 1 |
| 24 | Atk Helo Bn (24xOH58D) | 2.1 | 1 |
| 25 | Atk Helo Bn (24xAH64) | 4.5 | 1 |
| 26 | ADA Bn (Avenger) | 0.21 | 1 |
| 27 | Patriot Bn | 0.59 | 1 |
| 28 | Infantry Bn | 0.8 | 1 |
| 29 | Tank Co | 0.3 | 0.25 |
| 30 | LAV Co | 0.2 | 0.25 |
| 31 | AAV Co | 0.2 | 0.25 |
| 32 | FA Bn | 1.5 | 0.25 |
| 33 | AH-1 Co | 1 | 0.25 |
| 34 | MEF (Fwd) | 5.6 | 16 |

Table 4 outlines the initial combat values for the red side, where $y$ represents a platoon of type $y.type$, along with the explanatory text $y.text$. Here, $y.size = 1$ indicates that $y.combatv$ applies to the entire battalion [12, 13]. All of these are fixed parameters.

*Table 4: Combat values for the red side.*

| y.type | y.text | y.combatv | y.size |
|---|---|---|---|
| 1 | Infantry Bn (BTR-50 / 60) | 0.29 | 1 |
| 2 | Infantry Bn (BTR-70 / 80) | 0.36 | 1 |
| 3 | Infantry Bn (BMP-1 / 2) | 0.51 | 1 |
| 4 | Infantry Bn (BMP-3) | 0.65 | 1 |
| 5 | Infantry Bn (Light / Air Assault) | 0.35 | 1 |
| 6 | Infantry Bn (Airborne) | 0.5 | 1 |
| 7 | Recon Bn | 0.2 | 1 |
| 8 | AT Bn | 0.45 | 1 |
| 9 | AT Bn (AT Bde / Div) | 0.45 | 1 |
| 10 | AT Bn (IMIBn / AT Regt) | 0.5 | 1 |
| 11 | Tank Bn (MIB 40xT55) | 0.77 | 1 |
| 12 | Tank Bn (MIB 40xT62) | 0.77 | 1 |
| 13 | Tank Bn (MIB 40xT64 / T72) | 0.89 | 1 |
| 14 | Tank Bn (MIB 40xT80) | 1 | 1 |
| 15 | Tank Bn (TR 31xT55 / T62) | 0.6 | 1 |
| 16 | Tank Bn (TR 31xT64 / T72) | 0.69 | 1 |
| 17 | Tank Bn (TR 31xT80) | 0.78 | 1 |
| 18 | Indep Tank Bn (51xT55) | 0.98 | 1 |
| 19 | Indep Tank Bn (51xT62) | 0.98 | 1 |
| 20 | Indep Tank Bn (51xT64 / T72) | 1.13 | 1 |
| 21 | Indep Tank Bn (51xT80) | 1.28 | 1 |
| 22 | 2A36 Bn | 0.75 | 1 |
| 23 | 2A65 Bn | 0.75 | 1 |
| 24 | 2S1 Bn | 0.9 | 1 |
| 25 | 2S3 Bn | 1.05 | 1 |
| 26 | 2S4 Bn | 0.45 | 1 |
| 27 | 2S5 Bn | 1.13 | 1 |
| 28 | 2S7 Bn | 1.28 | 1 |
| 29 | 2S9 Bn | 0.6 | 1 |
| 30 | 2S19 / 23 Bn | 1.35 | 1 |
| 31 | 9A52 Bn | 3.6 | 1 |
| 32 | BM 21 Bn | 3.15 | 1 |
| 33 | BM 21V Bn | 1.04 | 1 |
| 34 | 9P140 Bn | 3.6 | 1 |
| 35 | BM 24 Bn | 3.6 | 1 |
| 36 | D20 Bn | 0.77 | 1 |
| 37 | D30 Bn | 0.63 | 1 |
| 38 | FROG Bn | 0.22 | 1 |
| 39 | M46 Bn | 0.78 | 1 |
| 40 | M240 Bn | 0.4 | 1 |
| 41 | SCUD Bn | 0.8 | 1 |
| 42 | SCUD-B Bn | 0.4 | 1 |
| 43 | SS-21 Bn | 0.6 | 1 |
| 44 | 9A51 Bn | 3.78 | 1 |
| 45 | Hind- D Bn | 3.33 | 1 |
| 46 | HOKUM / HAVOK Bn | 5.53 | 1 |
| 47 | Hind-E Bn | 4.17 | 1 |
| 48 | SA-4 Bn | 0.46 | 1 |
| 49 | SA-6 / 8 Bn | 0.11 | 1 |
| 50 | SA-11 / 12 Bn | 0.54 | 1 |
| 51 | SA-17 Bn | 0.76 | 1 |
| 52 | S-60 Bn | 0.34 | 1 |

## APPENDIX: LOSSES

Table 5 shows blue and red losses based on historical data across various combat values and scenarios [12, 13].

*Table 5: Blue and red losses based on different combat values and types.*

| F to E force ratio | 1:4 | | 1:3 | | 1:2 | | 1:1 | | 2:1 | | 3:1 | | 4:1 | |
|---|---|---|---|---|---|---|---|---|---|---|---|---|---|---|
| Friendly vs Enemy | 0.25 | | 0.33 | | 0.5 | | 1 | | 2 | | 3 | | 4 | |
| Deliberate Attack vs Deliberate Defense | 60% | 10% | 30% | 15% | 20% | 20% | 40% | 15% | 20% | 20% | 15% | 30% | 10% | 60% |
| Deliberate Attack vs Hasty Defense | 85% | 5% | 60% | 5% | 50% | 10% | 25% | 25% | 10% | 50% | 5% | 60% | 5% | 85% |
| Deliberate Defense vs Deliberate Attack | 60% | 10% | 30% | 15% | 20% | 20% | 15% | 40% | 10% | 65% | 10% | 30% | 10% | 60% |
| Deliberate Defense vs Hasty Attack | 50% | 20% | 40% | 25% | 30% | 25% | 25% | 50% | 10% | 50% | 10% | 35% | 20% | 50% |
| Hasty Attack vs Deliberate Defense | 50% | 20% | 35% | 10% | 50% | 10% | 50% | 25% | 25% | 30% | 25% | 40% | 20% | 50% |
| Hasty Attack vs Hasty Defense | 60% | 10% | 40% | 10% | 30% | 15% | 15% | 15% | 15% | 30% | 10% | 50% | 10% | 60% |
| Hasty Defense vs Deliberate Attack | 85% | 5% | 60% | 5% | 50% | 10% | 25% | 25% | 10% | 50% | 15% | 50% | 5% | 85% |
| Hasty Defense vs Hasty Attack | 60% | 10% | 50% | 10% | 30% | 15% | 15% | 15% | 15% | 30% | 10% | 40% | 10% | 60% |
| Meeting Engagement vs Meeting Engagement | 85% | 50% | 60% | 10% | 35% | 15% | 10% | 10% | 15% | 35% | 10% | 60% | 50% | 85% |
| | Blue | Red | Blue | Red | Blue | Red | Blue | Red | Blue | Red | Blue | Red | Blue | Red |

## APPENDIX: ADVANCE RATE

Table 6 presents advance rates based on historical data for various forces across different scenarios [15, 16].

*Table 6: Advance Rate in km per Day.*

| Row | STANDARD (UNMODIFIED) ADVANCE RATES | | | | |
|---|---|---|---|---|---|
| | Rates in km/day | | | | |
| | | Armored Division | Mechzd. Division | Infantry Division or Force | Horse Cavalry Division or Force |
| | | Column | | | |
| | | 1 | 2 | 3 | 4 |
| | **Against Intense Resistance** (P/P: 1.0-1.10) | | | | |
| 1 | Hasty defense/delay | 4.0 | 4.0 | 4.0 | 3.0 |
| 2 | Prepared defense | 2.0 | 2.0 | 2.0 | 1.6 |
| 3 | Fortified defense | 1.0 | 1.0 | 1.0 | 0.6 |
| | **Against Strong/Intense Resistance** (P/P: 1.11–1.25) | | | | |
| 4 | Hasty defense/delay | 5.0 | 4.5 | 4.5 | 3.5 |
| 5 | Prepared defense | 2.25 | 2.25 | 2.25 | 1.5 |
| 6 | Fortified defense | 1.25 | 1.25 | 1.25 | 0.7 |
| | **Against Strong Defense** (P/P: 1.26–1.45) | | | | |
| 7 | Hasty defense/delay | 6.0 | 5.0 | 5.0 | 4.0 |
| 8 | Prepared defense | 2.5 | 2.5 | 2.5 | 2.0 |
| 9 | Fortified defense | 1.5 | 1.5 | 1.5 | 0.8 |
| | **Against Moderate/Strong Resistance** (P/P: 1.46–1.75) | | | | |
| 10 | Hasty defense | 9.0 | 7.5 | 6.5 | 6.0 |
| 11 | Prepared defense | 4.0 | 3.5 | 3.0 | 2.5 |
| 12 | Fortified defense | 2.0 | 2.0 | 1.75 | 0.9 |
| | **Against Moderate Resistance** (P/P: 1.76–2.25) | | | | |
| 13 | Hasty defense/delay | 12.0 | 10.0 | 8.0 | 8.0 |
| 14 | Prepared defense | 6.0 | 5.0 | 4.0 | 3.0 |
| 15 | Fortified defense | 3.0 | 2.5 | 2.0 | 1.0 |
| | **Against Slight/Moderate Resistance** (P/P: 2.26–3.0) | | | | |
| 16 | Hasty defense/delay | 16.0 | 13.0 | 10.0 | 12.0 |
| 17 | Prepared defense | 8.0 | 7.0 | 5.0 | 6.0 |
| 18 | Fortified defense | 4.0 | 3.0 | 2.5 | 2.0 |
| | **Against Slight Resistance** (P/P: 3.01–4.25) | | | | |
| 19 | Hasty defense/delay | 20.0 | 16.0 | 12.0 | 15.0 |
| 20 | Prepared defense | 10.0 | 8.0 | 6.0 | 7.0 |

| 21 | Fortified defense | 5.0 | 4.0 | 3.0 | 4.0 |
| --- | --- | --- | --- | --- | --- |
| | **Against Negligible/Slight Resistance** (P/P: 4.26–6.00) | | | | |
| 22 | Hasty defense/delay | 40.0 | 30.0 | 18.0 | 28.0 |
| 23 | Prepared defense | 20.0 | 16.0 | 10.0 | 14.0 |
| 24 | Fortified defense | 10.0 | 8.0 | 6.0 | 7.0 |
| | **Against Negligible Resistance** (P/P: 6.00 plus) | | | | |
| 25 | Hasty defense/delay | 60.0 | 48.0 | 24.0 | 40.0 |
| 26 | Prepared defense | 30.0 | 24.0 | 12.0 | 12.0 |
| 27 | Fortified defense | 30.0 | 24.0 | 12.0 | 12.0 |